\documentclass[conference]{IEEEtran}
\IEEEoverridecommandlockouts

\usepackage{cite}
\usepackage{amsmath,amssymb,amsfonts}
\usepackage{algorithmic}
\usepackage{graphicx}
\usepackage{textcomp}
\usepackage{xcolor}
\usepackage{booktabs}
\usepackage[numbers]{natbib}
\usepackage{multirow}
\usepackage{tabularx}
\usepackage{authblk}

\def\BibTeX{{\rm B\kern-.05em{\sc i\kern-.025em b}\kern-.08em
    T\kern-.1667em\lower.7ex\hbox{E}\kern-.125emX}}
\begin{document}

\title{DA-LIF: Dual Adaptive Leaky Integrate-and-Fire Model for Deep Spiking Neural Networks}

\author[1]{Tianqing Zhang\thanks{$\dagger$ These authors contributed equally to this work and should be considered co-first authors.}$\dagger$}
\author[2]{Kairong Yu$\dagger$}
\author[1]{Jian Zhang}
\author[1,2]{Hongwei Wang\thanks{* Corresponding author.}*}

\affil[1]{College of Computer Science and Technology, Zhejiang University, Hangzhou, China}
\affil[2]{ZJU-UIUC Institute, Zhejiang University, Haining, China}

\affil[ ]{\texttt{\{zhangtianqing,jianzhang.22\}@zju.edu.cn}}
\affil[ ]{\texttt{\{kairong.22,hongweiwang\}@intl.zju.edu.cn}}

\maketitle

\begin{abstract}
Spiking Neural Networks (SNNs) are valued for their ability to process spatio-temporal information efficiently, offering biological plausibility, low energy consumption, and compatibility with neuromorphic hardware. However, the commonly used Leaky Integrate-and-Fire (LIF) model overlooks neuron heterogeneity and independently processes spatial and temporal information, limiting the expressive power of SNNs. In this paper, we propose the Dual Adaptive Leaky Integrate-and-Fire (DA-LIF) model, which introduces spatial and temporal tuning with independently learnable decays. Evaluations on both static (CIFAR10/100, ImageNet) and neuromorphic datasets (CIFAR10-DVS, DVS128 Gesture) demonstrate superior accuracy with fewer timesteps compared to state-of-the-art methods. Importantly, DA-LIF achieves these improvements with minimal additional parameters, maintaining low energy consumption. Extensive ablation studies further highlight the robustness and effectiveness of the DA-LIF model.
\end{abstract}

\begin{IEEEkeywords}
Spiking Neural Networks, Neuromorphic Computing, Brain-inspired Learning
\end{IEEEkeywords}

\section{Introduction}\label{sec:intro}

\begin{figure*}[h]
    \centering
    \includegraphics[width=\textwidth]{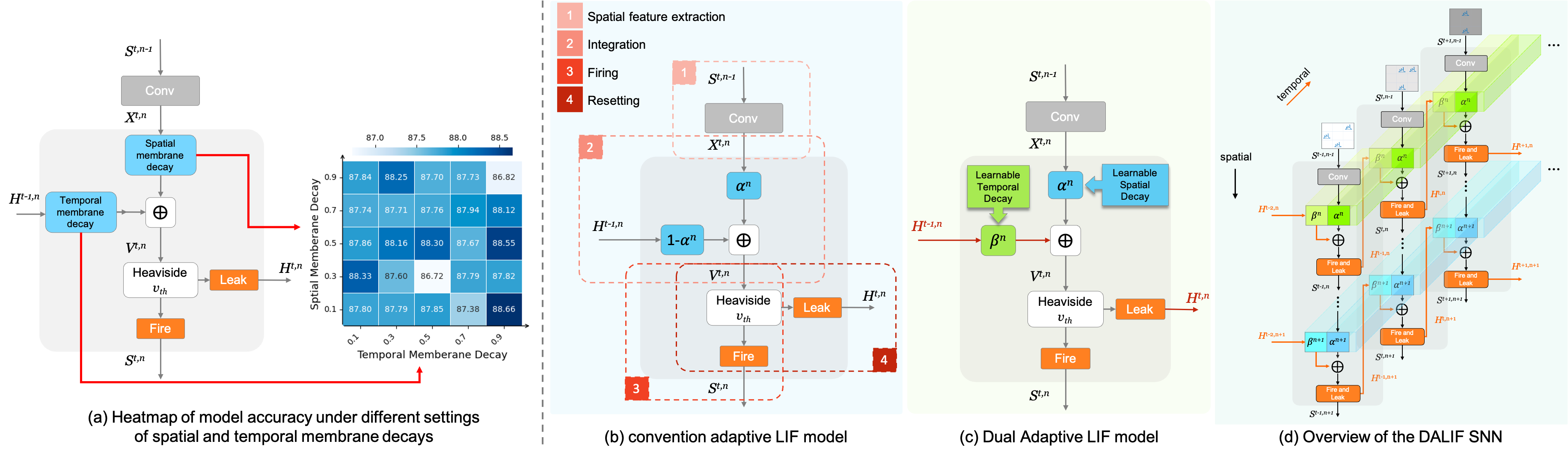}
    \caption{Overview of the DA-LIF model and DA-LIF-SNN Architecture.
    (a) Experimental Motivation of the proposed method.
    (b) convention adaptive LIF model
    (c) proposed DA-LIF model
    (d) \(\alpha^n\) and \(\beta^n\) are independently learned and share parameters within the same layer, but differ across each layer in networks.
}
    \label{fig:overview}
\end{figure*}

The human brain’s efficiency and complexity have long inspired advancements in artificial intelligence, particularly in deep learning and Convolutional Neural Networks (CNNs). Despite CNNs’ success, their energy consumption remains much higher than that of the human brain. Spiking Neural Networks (SNNs), modeled more closely on biological neural activity, have emerged as the third generation of neural networks, offering better biological plausibility, lower energy use, and compatibility with neuromorphic hardware~\cite{maass_networks_1997}.
SNNs simulate how neurons communicate via spikes, but current models often overlook the heterogeneity and complex dynamics of biological neurons. Insights from neuroscience highlight the importance of factors like neurotransmitter clearance and modulation of synaptic strength. These mechanisms, while critical in biological systems, are not yet fully integrated into SNNs.
To bridge this gap, we propose incorporating glial cell functions into the Leaky Integrate-and-Fire (LIF) neuron model by making membrane decay parameters adaptive\cite{turrigiano_homeostatic_2004,alcami_beyond_2019,abbott_lapicques_1999}.
The experimental results in Fig.~\ref{fig:overview}(a) demonstrate that both spatial and temporal membrane decays contribute to the network accuracy, with a maximum improvement of 1.84\%. 
To optimizing the values for two decay settings automatically, we advocate the incorporation of independently learnable membrane decays in LIF neurons to regulate their membrane permeability, thereby achieving adaptive learning and allowing relevant information to pass through while blocking irrelevant information without imposing a significant computational burden. 
The strategic incorpration introduces intricate biological mechanisms to the SNNs with few additional parameters, enhances the fidelity of SNNs, paving the way for innovative solutions in neural network modeling.
Our contributions summarized as follows:
\begin{itemize}
    \item 
    Incorporation of independently learnable spatial and temporal decays, enabling more precise regulation of membrane permeability across layers, improving performance without compromising efficiency.
    \item 
    We evaluate our method on both static and neuromorphic datasets, outperforming previous state-of-the-art results with fewer timesteps.
    \item 
    Extensive ablation experiments demonstrate the robustness and effectiveness of the proposed model.
\end{itemize}

\section{Related Works}
Training of SNN typically follows two approaches: Artificial Neural Network (ANN)-to-SNN conversion and direct SNN training. The ANN-to-SNN method involves converting pre-trained ANNs by mapping parameters and replacing ReLU activation with spiking activation. 
For direct SNN training, recent advancements have introduced various improvements. The tdBN \cite{zheng_going_2021} enhances feature normalization in temporal and spatial dimensions , while MPBN \cite{guo_membrane_2023} and TET stabilize the training process through better batch normalization and momentum control. 
Other innovations, such as SEW-ResNet and DS-ResNet, improve the adaptation of ResNet architectures to SNNs\cite{fang_deep_2021,feng_multi-level_2022,hu_advancing_2024}. 
Techniques like IM-Loss \cite{guo_im-loss_2022} and model compression strategies also optimize information processing and advancing the applicability of SNNs\cite{xu_constructing_2023}.

Several studies have demonstrated the advantages of adaptive LIF models\cite{chen_gradual_2022,liu_event-based_2022}. For example, LSNN \cite{bellec_long_2018} and LTMD\cite{wang_ltmd_2022} introduced adaptive threshold neurons, improving the learning dynamics of SNNs. Additionally, works such as PLIF incorporated a learnable membrane time constant to enhance the performance of spiking neurons\cite{fang_incorporating_2021}. Recent efforts like Diet-SNN and BDETT have further optimized neuron models by integrating learnable parameters such as membrane leak and firing thresholds\cite{rathi_diet-snn_2023,ding_biologically_2022,xu2024rsnn}.
Despite these advances, current models do not independently address spatial and temporal aspects of neuron behavior. Future work could focus on separate learnable parameters for each dimension and layer, aligning more closely with biological observations and enhancing SNN expressiveness.

\section{Methods}

\subsection{LIF model with Variable Decays}\label{sec:1}

Spiking neurons serve as the primary computational units in SNNs. 
Among various neuron models, the LIF model commonly used in deep SNNs is expressed by:

\begin{equation}
    \tau_\text{m} \frac{du(t)}{dt} = -(u(t) - u_\text{rest}) + R_m I(t),
\end{equation}
where \(\tau_\text{m}\) is the membrane time constant, \(u(t)\) is the membrane potential, \(u_\text{rest}\) is the resting potential, \(I(t)\) is the pre-synaptic input, and $R_m$ is the membrane resistance.
When $u(t)$ exceeds a threshold $u_{th}$, the neuron fires a spike and $u(t)$ is reset. 

To adapt the LIF model for deep learning, discretize it into iterative equations:

\begin{subequations}
    \begin{align}
     \quad & V^{t,n} = f(H^{t-1,n}, X^{t,n}), \label{equ:charging} \\
     \quad & S^{t,n} = \Theta(V^{t,n} - v_{\text{th}}), \label{equ:firing}\\
     \quad & H^{t,n} = V_{\text{reset}} \cdot S^{t,n} + V^{t,n} \odot (1 - S^{t,n}). \label{equ:resetting}
\end{align}
\end{subequations}
Where \(V^{t,n}\) is the membrane potential after integration, 
\(H^{t-1,n}\) represents the membrane potential after a spike trigger in the previous timestep,
The input current \(X^{t,n}\) at timestep \(t\) layer \(n\) is obtained through the output spiking of the previous layer \(S^{t,n-1}\) via a convolutional layer with weight \(W\):
\begin{equation}\label{equ:con}
    X^{t,n} = \text{Conv}(W,S^{t,n-1}).
\end{equation}
The function \(f(\cdot)\) for the vanilla LIF model is:
\begin{equation}\label{equ:clear}
    V^{t,n} = H^{t-1,n} + \frac{1}{\tau_\text{m}}(X^{t,n} + (- (H^{t-1,n} - V_{\text{rest}}))),
\end{equation}
The Heaviside step function \(\Theta(x)\) return 0 if \(x<0\), 1 otherwise.
The subsequent membrane potential \(H^{t,n}\) reset to \(V_{\text{reset}}\) if \(S^{t,n} = 1\) and maintain \(V^{t,n}\) otherwise.
In this study, \(V_{\text{rest}}\) and \(V_{\text{reset}}\) are both set to 0.

To enhance adaptability, variable decay terms \(\mu\) and \(\lambda\) are introduced into the LIF model, allowing the membrane potential and input signal to decay independently:
\begin{equation}\label{equ:mu}
    \tau_\text{m} \frac{du(t)}{dt} = -\mu(u(t) - u_\text{rest}) + \lambda I(t).
\end{equation}
Then the integration of the LIF model with variable decays is:
\begin{equation}\label{equ:clear}
    V^{t,n} = (1-\frac{\mu}{\tau_\text{m}})H^{t-1,n} + \frac{\lambda}{\tau_\text{m}} X^{t,n}.
\end{equation}

\subsection{LIF with Dual Adaptive Mechanism}\label{sec:2}

Assigning independent learning of temporal and spatial decay parameters by employing \(1-\frac{\mu}{\tau_\text{m}}\) and \(\frac{\lambda}{\tau_\text{m}}\) as distinct variable decay terms.
Firstly, define: 
\begin{equation}
    \alpha' = \frac{\lambda}{\tau_\text{m}}, \quad
    \beta' = 1 - \frac{\mu}{\tau_\text{m}}.
\end{equation}
The conflict arises when \(\alpha'\) and \(\beta'\)are learnable decays with a shared parameter \(\tau_\text{m}\), whereas our method aims to train the two learnable decays independently.
decouple the spatial and temporal learnable decays, which requires the two learnable decays are trained independently.
Then by defining
\begin{equation}
    \alpha = \frac{\lambda}{\tau_\text{a}}, \quad
    \beta = 1 - \frac{\mu}{\tau_\text{b}},
\end{equation}
where \(\tau_\text{a}\) and \(\tau_\text{b}\) are separate time constants for spatial and temporal processes. The membrane potential update becomes:
\begin{equation}\label{equ:BN}
    V^{t,n} = \beta^n H^{t-1,n} + \alpha^n X^{t,n}.
\end{equation}
The connection between the Dual Adaptive LIF neurons is depicted in Fig.~\ref{fig:overview}(c), where the spatial and temporal tuning, \(\alpha^n\) and \(\beta^n\), are independently learned and share parameters within the same layer but at different timesteps, with distinct parameters when trained across different layers. 

\subsection{Training Framework}\label{sec:3}

Addressing the challenge of the non-differentiable nature of spiking activity, we employ the spatio-temporal backpropagation (STBP) combined with surrogate gradients (SG)\cite{wu_spatio-temporal_2018}:

\begin{equation}
            \frac{\partial L}{\partial W^n} = 
            \sum \limits_t \left(
            \frac{\partial L}{\partial S^{t,n}} \cdot \frac{\partial S^{t,n}}{\partial V^{t,n}} + \frac{\partial L}{\partial V^{t+1,n}} \cdot \frac{\partial V^{t+1,n}}{\partial V^{t,n}}
             \right)
            \frac{\partial V^{t,n}}{\partial W^n}.
\end{equation}

\begin{equation}
    \frac{\partial S^t}{\partial V^t} = 
    \frac{1}{a} \cdot \text{sign}(|V^{t}-V_\text{th}|< \frac{a}{2}),
\end{equation}
where \(a\) is a hyper-parameter and set to 1, the gradient is equal to 1 when  \(V_\text{th}-0.5 \leq V^{t} \leq V_\text{th}+0.5\), and 0 otherwise. 

\section{Experiments}

\subsection{Dataset and Implementation Details}\label{sec:implementation}
\subsubsection{Dataset}
The CIFAR10 dataset consists of 50,000 training and 10,000 test images across 10 classes, while CIFAR100~\cite{krizhevsky_cifar-10_2010} extends it to 100 classes for more fine-grained classification. 
ImageNet~\cite{deng_imagenet_2009}, with 1.2 million images in 1,000 classes, is a widely used benchmark for large-scale image classification. 
The CIFAR10-DVS~\cite{li_cifar10-dvs_2017} dataset is one of the largest visual neuromorphic datasets, while the DVS128 Gesture~\cite{amir_low_2017} dataset is specifically ideal for gesture recognition, making them well-suited for evaluating SNNs.

\subsubsection{Implementation Details}

In our implementation, critical hyperparameters such as the firing threshold \(v_\text{th}\) are set to 1.0, with the learnable decay terms \(\alpha\) and \(\beta\) both initialized at 1.0 within the range of [-1, 1]. The codebase is fully implemented in PyTorch. All experiments, except those involving ImageNet, are conducted on an RTX 3090 GPU, while ImageNet experiments are executed on 8 RTX 4090D GPUs. Optimization is performed using the SGD optimizer with a momentum of 0.9, and the initial learning rate is set to 0.1. For CIFAR10, CIFAR100, and CIFAR10-DVS, models are trained for 1000 epochs, while the DVS CIFAR-10 dataset is trained for 200 epochs. On ImageNet, models are trained for 320 epochs.

\subsection{Comparison to Previous Work}

\begin{table*}
    \setlength{\tabcolsep}{2pt}
\centering
\caption{Comparison results with SOTA methods on Static Datasets.}
\begin{tabular}{cccccccccc}
    \toprule    
\multirow{2}{*}{\textbf{Method}}   & \multicolumn{3}{c}{\textbf{CIFAR-10}}            & \multicolumn{3}{c}{\textbf{CIFAR-100}}         & \multicolumn{3}{c}{\textbf{ImageNet}}   \\
\cmidrule{2-10}
                          & \textbf{Architecture} & \textbf{Timestep} & \textbf{Accuracy} & \textbf{Architecture} & \textbf{Timestep} & \textbf{Accuracy} & \textbf{Architecture} & \textbf{Timestep} & \textbf{Accuracy} \\
\midrule
RecDis-SNN\cite{guo_recdis-snn_2022}    & ResNet-19 & 4 & 95.53\% & ResNet-19 & 4 & 74.10\% & ResNet-34 & 6 & 67.33\% \\
\hline
\multirow{2}{*}{GLIF\cite{yao_glif_2022}}     & ResNet-19 & 2 & 94.44\% & ResNet-19 & 2 & 75.48\% & ResNet-34 & 4 & 67.52\% \\
                          & ResNet-19 & 4 & 94.85\% & ResNet-19 & 4 & 77.05\% & - & - & - \\
\hline
\multirow{2}{*}{TET\cite{deng_temporal_2021}} & ResNet-19 & 2 & 94.16\% & ResNet-19 & 2 & 72.87\% & ResNet-34 & 6 & 64.79\% \\
                          & ResNet-19 & 4 & 94.44\% & ResNet-19 & 4 & 74.47\% & - & - & - \\
\hline
\multirow{2}{*}{LSG\cite{lian_learnable_2023}} & ResNet-19 & 2 & 94.41\% & ResNet-19 & 2 & 76.32\% & - & - & - \\
                          & ResNet-19 & 4 & 95.17\% & ResNet-19 & 4 & 76.85\% & - & - & - \\
\hline
\multirow{2}{*}{PFA\cite{deng_tensor_2024}}  & ResNet-19 & 2 & 95.60\% & ResNet-19 & 2 & 76.70\% & - & - & - \\
                          & ResNet-19 & 4 & 95.71\% & ResNet-19 & 4 & 78.10\% & - & - & - \\
\hline
Diet-SNN\cite{rathi_diet-snn_2023} & ResNet-20 & 5 & 91.78\% & ResNet-20 & 5 & 64.07\% & - & - & - \\
\hline
\multirow{2}{*}{IM-loss\cite{guo_im-loss_2022}}  & ResNet-19 & 2 & 93.85\% & - & - & - & ResNet-18 & 6 & 67.43\% \\
                          & ResNet-19 & 4 & 95.40\% & - & - & - & - & - & - \\
\hline
\multirow{5}{*}{MPBN\cite{guo_membrane_2023}} & ResNet-19 & 2 & 96.47\% & ResNet-19 & 2 & 79.51\% & ResNet-18 & 4 & 63.14\% \\
                          & ResNet-19 & 4 & 96.52\% & ResNet-19 & 4 & 80.10\% & ResNet-34 & 4 & 64.71\% \\
                          & ResNet-20 & 2 & 93.54\% & ResNet-20 & 2 & 70.79\% & - & - & - \\
                          & ResNet-20 & 4 & 94.28\% & ResNet-20 & 4 & 72.30\% & - & - & - \\
                          & VGG-16 & 2 & 93.96\% & VGG-16 & 2 & 73.88\% & - & - & - \\
\hline
\multirow{2}{*}{IM-LIF\cite{lian_im-lif_2024}}  & ResNet-19 & 3 & 95.29\% & ResNet-19 & 3 & 77.21\% & - & - & - \\
                          & ResNet-19 & 6 & 95.66\% & ResNet-19 & 6 & 77.42\% & - & - & - \\      
\hline
\multirow{7}{*}{\textbf{Ours}}  & ResNet-19 & 1 & $\textbf{96.44\%} \pm 0.10\%$& ResNet-19 & 1 & \textbf{$79.77\% \pm 0.10\%$}& ResNet-18 & 4 & $\textbf{68.09\%} \pm 0.20\%$ \\
                          & ResNet-19 & 2 & $\textbf{96.55\%} \pm 0.07\%$& ResNet-19 & 2 & $\textbf{80.59\%} \pm 0.08\%$& ResNet-34 & 4 & $\textbf{70.58\%} \pm 0.20\%$ \\
                          & ResNet-19 & 4 & $\textbf{96.72\%} \pm 0.10\%$& ResNet-19 & 4 & $\textbf{80.16\%} \pm 0.11\%$& - & - & - \\
                          & ResNet-20 & 1 & $\textbf{92.89\%} \pm 0.12\%$& ResNet-20 & 1 & $\textbf{69.37\%} \pm 0.08\%$& - & - & - \\
                          & ResNet-20 & 2 & $\textbf{93.65\%} \pm 0.07\%$& ResNet-20 & 2 & $\textbf{72.07\%} \pm 0.10\%$& - & - & - \\
                          & ResNet-20 & 4 & $\textbf{94.16\%} \pm 0.08\%$& ResNet-20 & 4 & $\textbf{73.25\%} \pm 0.08\%$& - & - & - \\
                          & VGG-16 & 2 & $\textbf{94.55\%} \pm 0.10\%$ & VGG-16 & 2 & $\textbf{74.36\%} \pm 0.08\%$ & - & - & - \\                          
\bottomrule
\end{tabular}
\label{tab:exp_static}
\end{table*}

\begin{table}
\renewcommand{\arraystretch}{0.9} 
    \caption{Comparison Results with SOTA Methods on Neuromorphic Datasets.}
    \centering
    \begin{tabular}{llrr}
        \toprule
         \textbf{Methods} & \textbf{Architecture} & \textbf{Timestep} & \textbf{Accuracy} \\
        \midrule
        \multicolumn{4}{c}{\textbf{CIFAR10-DVS}}\\
        \midrule
         IM-loss~\cite{guo_im-loss_2022} & ResNet-19 & 10 & 72.60\% \\
         LSG~\cite{lian_learnable_2023} & ResNet-19 & 10 & 77.90\% \\
         MPBN~\cite{guo_membrane_2023} & ResNet-19 & 10 & 74.40\% \\
         MPBN~\cite{guo_membrane_2023} & ResNet-20 & 10 & 78.70\% \\
         TET~\cite{deng_temporal_2021} & VGGSNN & 10 & 77.30\% \\
         GLIF~\cite{yao_glif_2022} & 7B-wideNet & 16 & 78.10\% \\
         STSA~\cite{wang_spatial-temporal_2023} & STS-Transformer & 16 & 79.93\%\\
         PLIF~\cite{fang_incorporating_2021} & PLIFNet & 20 & 74.80\% \\
        \midrule
        \multirow{3}{*}{\textbf{Ours}} & ResNet-19 & 10 & $78.00\% \pm 0.20\%$ \\
         & ResNet-20 & 16 & $82.10\% \pm 0.20\%$ \\
         & VGG-16 & 16 & $\textbf{82.42\%} \pm 0.20\%$ \\
        \midrule
        \multicolumn{4}{c}{\textbf{DVS128 Gesture}}\\
        \midrule
         STBP-tdBN~\cite{zheng_going_2021} & ResNet-17 & 40 & 96.87\%\\
         SEW~\cite{fang_deep_2021} & 7B-Net & 16 & 97.92\% \\
         PLIF~\cite{fang_incorporating_2021} & PLIFNet & 20 & 97.57\% \\
         MA-SNN~\cite{yao_attention_2023} & 5 layers SCNN & 20 & 98.23\% \\
         ASA-SNN~\cite{yao_inherent_2023} & 5 layers SCNN & 20 &  97.70\%\\
         LIAF+TA\cite{yao_temporal-wise_2021} & TA-SNN-Net & 60 & 98.61\%\\
        \midrule
         \textbf{Ours} & VGG-11 & 16 & $\textbf{98.61\%} \pm 0.20\%$ \\
        \bottomrule
    \end{tabular}
    \label{tab:neuromorphic}
\end{table}

\paragraph{Static Image Classification}
As shown in Tab.~\ref{tab:exp_static}, the DA-LIF SNN model was tested on CIFAR-10/100 and ImageNet, showing superior accuracy of 96.72\%/80.59\% and 70.58\% respectively. 
Integrated into ResNet-19, ResNet-20, and VGG-16 architectures, and using timesteps of 1, 2, and 4, the model consistently outperforms SOTA methods. Notably, it achieved a 1.28\% accuracy improvement on CIFAR-100 with ResNet-20, timestep 2.

\paragraph{Event-based Action Recognition}
As shown in Tab.~\ref{tab:neuromorphic}, the DA-LIF achieving a 1.71\% improvement on CIFAR10-DVS and reaching 98.61\% accuracy on DVS128 Gesture with only 16 timesteps. 

\subsection{Analysis of Computation Efficiency}\label{sec:energy}

Tab.~\ref{tab:energy} shows the computational cost for processing a single sample from CIFAR-100, ImageNet, and CIFAR10-DVS using ResNet architectures. For ANNs, energy is calculated based on MAC operations, while for SNNs, energy depends on both MACs in the first layer and accumulate computations (ACs) in subsequent layers, influenced by spiking activity and timesteps\cite{kundu_recent_2024}. SNNs show greater energy efficiency, with each synaptic operation in an ANN equating to roughly five in an SNN. This highlights the potential for substantial energy savings in SNNs, especially with optimized hardware designs.

\begin{table}
\setlength{\tabcolsep}{2pt} 
\renewcommand{\arraystretch}{0.9} 
    \centering
    \caption{Computational consumption for processing a single sample.}
    \begin{tabular}{lrrrrrrrrr}
    \toprule
    \textbf{Dataset} & \textbf{Timestep} & \textbf{ACs} & \textbf{MACs} & \textbf{FLOPs} & \textbf{Params} & \textbf{Energy} \\
    \midrule
    CIFAR-100         & 4      & 143.31M    & 56.55M        & 873.47M      & 11.3M & 0.3891mJ\\
    \midrule
    ImageNet        & 4      & 1.33G      & 974.86M       & 7.33G        & 11.7M & 5.6814mJ\\
    \midrule
    CIFAR10-DVS      & 16     & 1.42G      & 863.85M       & 13.93G       & 11.2M & 5.2517mJ\\
    \bottomrule
    \end{tabular}
    \label{tab:energy}
\end{table}

\subsection{Ablation Study}\label{sec:abla}
\begin{figure}
    \centering
    \includegraphics[width=0.9\columnwidth]{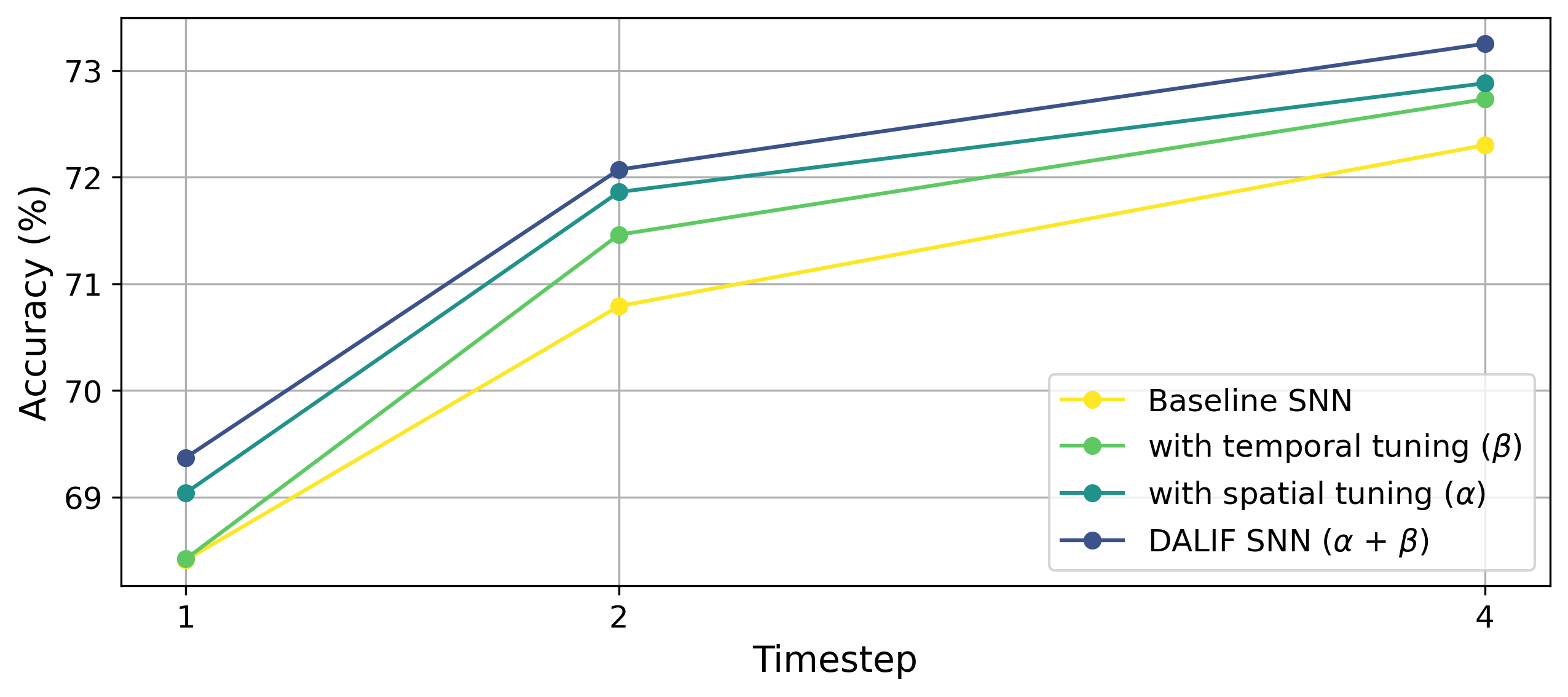}
    \caption{Impact of \(\alpha\) and \(\beta\) on CIFAR-100 with ResNet-20}
    \label{fig:abal_plot}
\end{figure}

\begin{figure}
    \centering
    \includegraphics[width=0.9\columnwidth]{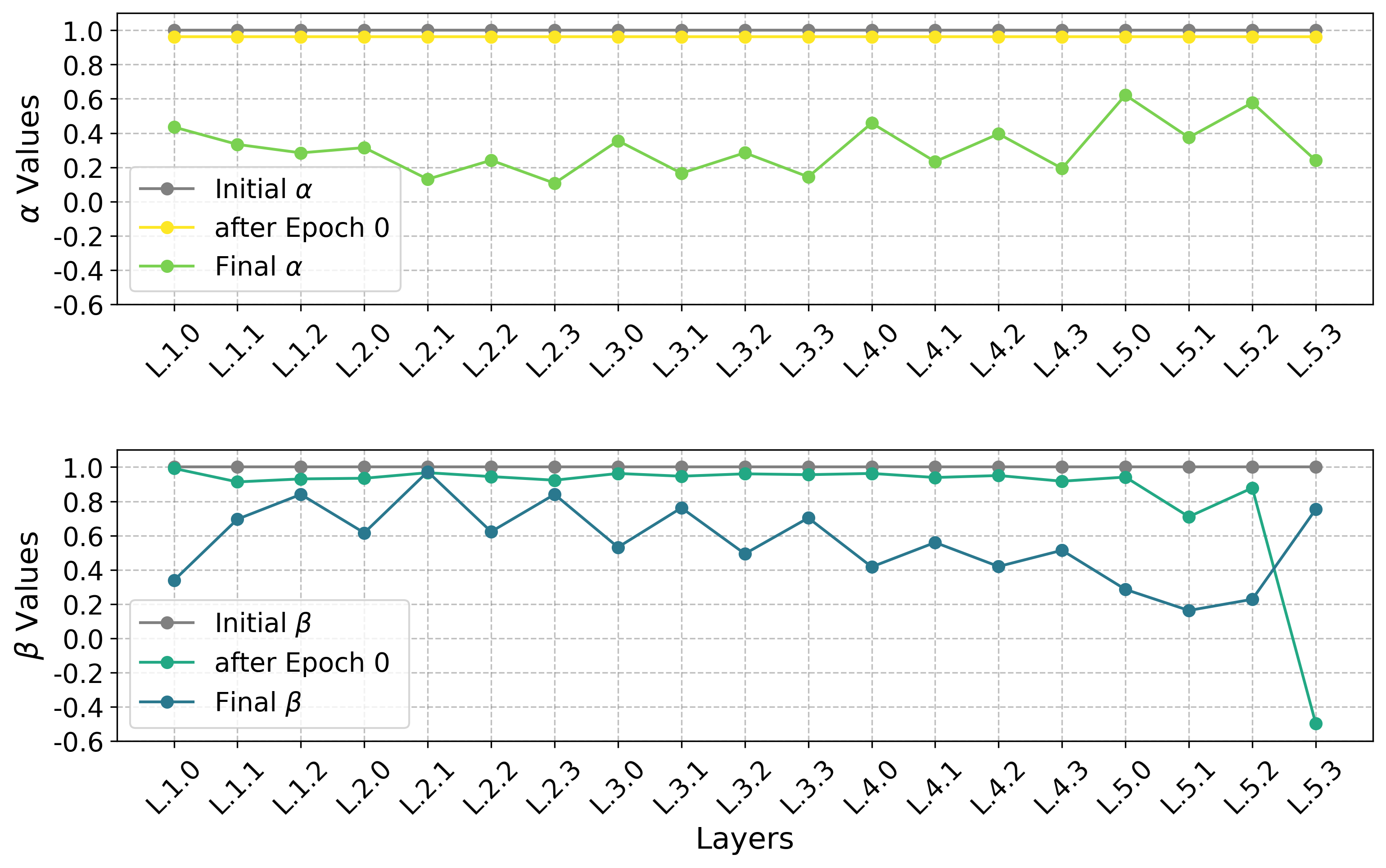}
    \caption{Distribution of $\alpha$ and $\beta$ Across Layers.}
    \label{fig:abla1}
\end{figure}
\subsubsection{Impact of Spatial and Temporal Tuning}

We evaluated the individual contributions of the \(\alpha\) and \(\beta\) in DA-LIF neurons by training each decay independently on the CIFAR-100 dataset using ResNet-20. As depicted in Fig.~\ref{fig:abal_plot}, tuning \(\alpha\) or \(\beta\) alone improved accuracy by 0.76\% and 0.37\% over the baseline, respectively. 
At timestep = 1, \(\beta\) had no effect, as it matched the baseline SNN. 
The combined tuning of both \(\alpha\) and \(\beta\) led to a 1.28\% accuracy increase, highlighting the enhanced feature extraction ability provided by learnable decays in both spatial and temporal dimensions.

\subsubsection{Evaluation of Different Activation Functions.}

The impact of different activation functions, $Sigmoid$ and $Tanh$ were evaluated on the network’s performance across two datasets, CIFAR-10 and CIFAR-100. 
The parameters for these functions were initialized to 1. 
Using the ResNet-20 architecture with a timestep of 4. 
The findings indicate that $Tanh$ outperforms $Sigmoid$, improving performance by 0.48\% on CIFAR-10 and 0.27\% on CIFAR-100. 
Given that the $Tanh$ function constrains values within the range of [-1, 1], it is the preferred choice for the proposed model due to its superior performance.

\subsubsection{Distribution of $\alpha$ and $\beta$ Across Layers.} 

The distribution of \(\alpha\) and \(\beta\) values across the layers of the neural network during training is shown in Fig.~\ref{fig:abla1}, reveals three key insights.
Firstly, the significant gap between the initial and final values demonstrate the effectiveness of the training process.
Second, distinct  \(\alpha\) and \(\beta\) values across layers indicate adaptability, with layers focusing on different features.
Finally, shallow layers prioritize temporal features, while deeper layers emphasize spatial features, highlighting the dynamic spatio-temporal tuning necessary for improved network performance.

\section{Conclusion}
The DA-LIF model is a novel approach based on investigating the role of glial cells in neural synapses and supported by experimental findings.
It introduces spatial and temporal tuning mechanisms that regulate neuron membrane permeability, allowing selective feature extraction from input signals in both spatial and temporal dimensions. By incorporating independently learnable decay parameters across neurons and layers, the model achieves heterogeneous filtering and prioritization of inputs. 
This results in superior performance, achieving state-of-the-art accuracy on static and neuromorphic datasets with fewer timesteps. Extensive ablation studies further validate the model’s robustness and effectiveness.

\section*{Acknowledgment}

This work was supported in part by Zhejiang Provincial Natural Science Foundation of China (LDT23F02023F02).

\bibliographystyle{IEEEtran}
\small
\bibliography{references}

\begin{thebibliography}{10}
\providecommand{\url}[1]{#1}
\csname url@samestyle\endcsname
\providecommand{\newblock}{\relax}
\providecommand{\bibinfo}[2]{#2}
\providecommand{\BIBentrySTDinterwordspacing}{\spaceskip=0pt\relax}
\providecommand{\BIBentryALTinterwordstretchfactor}{4}
\providecommand{\BIBentryALTinterwordspacing}{\spaceskip=\fontdimen2\font plus
\BIBentryALTinterwordstretchfactor\fontdimen3\font minus \fontdimen4\font\relax}
\providecommand{\BIBforeignlanguage}[2]{{%
\expandafter\ifx\csname l@#1\endcsname\relax
\typeout{** WARNING: IEEEtran.bst: No hyphenation pattern has been}%
\typeout{** loaded for the language `#1'. Using the pattern for}%
\typeout{** the default language instead.}%
\else
\language=\csname l@#1\endcsname
\fi
#2}}
\providecommand{\BIBdecl}{\relax}
\BIBdecl

\bibitem{maass_networks_1997}
W.~Maass, ``Networks of spiking neurons: {The} third generation of neural network models,'' \emph{Neural Networks}, vol.~10, no.~9, pp. 1659--1671, Dec. 1997.

\bibitem{turrigiano_homeostatic_2004}
G.~G. Turrigiano and S.~B. Nelson, ``\BIBforeignlanguage{en}{Homeostatic plasticity in the developing nervous system},'' \emph{\BIBforeignlanguage{en}{Nature Reviews Neuroscience}}, vol.~5, no.~2, pp. 97--107, Feb. 2004.

\bibitem{alcami_beyond_2019}
P.~Alcamí and A.~E. Pereda, ``\BIBforeignlanguage{en}{Beyond plasticity: the dynamic impact of electrical synapses on neural circuits},'' \emph{\BIBforeignlanguage{en}{Nature Reviews Neuroscience}}, vol.~20, no.~5, pp. 253--271, May 2019.

\bibitem{abbott_lapicques_1999}
L.~F. Abbott, ``\BIBforeignlanguage{en}{Lapicque’s introduction of the integrate-and-fire model neuron (1907)},'' \emph{\BIBforeignlanguage{en}{Brain Research Bulletin}}, vol.~50, no. 5-6, pp. 303--304, Nov. 1999.

\bibitem{zheng_going_2021}
H.~Zheng, Y.~Wu, L.~Deng, Y.~Hu, and G.~Li, ``\BIBforeignlanguage{en}{Going {Deeper} {With} {Directly}-{Trained} {Larger} {Spiking} {Neural} {Networks}},'' \emph{\BIBforeignlanguage{en}{Proceedings of AAAI}}, vol.~35, pp. 11\,062--11\,070, May 2021, number: 12.

\bibitem{guo_membrane_2023}
Y.~Guo, Y.~Zhang, Y.~Chen, W.~Peng, X.~Liu, L.~Zhang, X.~Huang, and Z.~Ma, ``Membrane {Potential} {Batch} {Normalization} for {Spiking} {Neural} {Networks},'' in \emph{Proceedings of ICCV}, Oct. 2023, pp. 19\,420--19\,430.

\bibitem{fang_deep_2021}
W.~Fang, Z.~Yu, Y.~Chen, T.~Huang, T.~Masquelier, and Y.~Tian, ``Deep {Residual} {Learning} in {Spiking} {Neural} {Networks},'' \emph{Proceedings of NeurIPS}, vol.~34, pp. 21\,056--21\,069, 2021.

\bibitem{feng_multi-level_2022}
L.~Feng, Q.~Liu, H.~Tang, D.~Ma, and G.~Pan, ``\BIBforeignlanguage{en}{Multi-{Level} {Firing} with {Spiking} {DS}-{ResNet}: {Enabling} {Better} and {Deeper} {Directly}-{Trained} {Spiking} {Neural} {Networks}},'' in \emph{\BIBforeignlanguage{en}{Proceedings of IJCAI}}, Jul. 2022, pp. 2471--2477.

\bibitem{hu_advancing_2024}
Y.~Hu, L.~Deng, Y.~Wu, M.~Yao, and G.~Li, ``Advancing {Spiking} {Neural} {Networks} {Toward} {Deep} {Residual} {Learning},'' \emph{TNNLS}, pp. 1--15, 2024.

\bibitem{guo_im-loss_2022}
Y.~Guo, Y.~Chen, L.~Zhang, X.~Liu, Y.~Wang, X.~Huang, and Z.~Ma, ``\BIBforeignlanguage{en}{{IM}-{Loss}: {Information} {Maximization} {Loss} for {Spiking} {Neural} {Networks}},'' \emph{\BIBforeignlanguage{en}{Proceedings of NeurIPS}}, vol.~35, pp. 156--166, Dec. 2022.

\bibitem{xu_constructing_2023}
Q.~Xu, Y.~Li, J.~Shen, J.~K. Liu, H.~Tang, and G.~Pan, ``\BIBforeignlanguage{en}{Constructing {Deep} {Spiking} {Neural} {Networks} {From} {Artificial} {Neural} {Networks} {With} {Knowledge} {Distillation}},'' in \emph{\BIBforeignlanguage{en}{Proceedings of {CVPR}}}, Jun. 2023, pp. 7886--7895.

\bibitem{chen_gradual_2022}
Y.~Chen, S.~Zhang, S.~Ren, and H.~Qu, ``Gradual {Surrogate} {Gradient} {Learning} in {Deep} {Spiking} {Neural} {Networks},'' in \emph{ICASSP}, May 2022, pp. 8927--8931, iSSN: 2379-190X.

\bibitem{liu_event-based_2022}
Q.~Liu, D.~Xing, L.~Feng, H.~Tang, and G.~Pan, ``Event-{Based} {Multimodal} {Spiking} {Neural} {Network} with {Attention} {Mechanism},'' in \emph{ICASSP}, May 2022, pp. 8922--8926, iSSN: 2379-190X.

\bibitem{bellec_long_2018}
G.~Bellec, D.~Salaj, A.~Subramoney, R.~Legenstein, and W.~Maass, ``\BIBforeignlanguage{en}{Long short-term memory and {Learning}-to-learn in networks of spiking neurons},'' \emph{\BIBforeignlanguage{en}{Proceedings of NeurIPS}}, vol.~31, 2018.

\bibitem{wang_ltmd_2022}
S.~Wang, T.~H. Cheng, and M.-H. Lim, ``\BIBforeignlanguage{en}{{LTMD}: {Learning} {Improvement} of {Spiking} {Neural} {Networks} with {Learnable} {Thresholding} {Neurons} and {Moderate} {Dropout}},'' \emph{\BIBforeignlanguage{en}{Proceedings of NeurIPS}}, vol.~35, pp. 28\,350--28\,362, Dec. 2022.

\bibitem{fang_incorporating_2021}
W.~Fang, Z.~Yu, Y.~Chen, T.~Masquelier, T.~Huang, and Y.~Tian, ``\BIBforeignlanguage{en}{Incorporating {Learnable} {Membrane} {Time} {Constant} {To} {Enhance} {Learning} of {Spiking} {Neural} {Networks}},'' in \emph{\BIBforeignlanguage{en}{Proceedings of {ICCV}}}, 2021, pp. 2661--2671.

\bibitem{rathi_diet-snn_2023}
N.~Rathi and K.~Roy, ``{DIET}-{SNN}: {A} {Low}-{Latency} {Spiking} {Neural} {Network} {With} {Direct} {Input} {Encoding} and {Leakage} and {Threshold} {Optimization},'' \emph{IEEE TNNLS}, vol.~34, no.~6, pp. 3174--3182, Jun. 2023.

\bibitem{ding_biologically_2022}
J.~Ding, B.~Dong, F.~Heide, Y.~Ding, Y.~Zhou, B.~Yin, and X.~Yang, ``\BIBforeignlanguage{en}{Biologically {Inspired} {Dynamic} {Thresholds} for {Spiking} {Neural} {Networks}},'' \emph{\BIBforeignlanguage{en}{Proceedings of NeurIPS}}, vol.~35, pp. 6090--6103, Dec. 2022.

\bibitem{xu2024rsnn}
Q.~Xu, X.~Fang, Y.~Li, J.~Shen, D.~Ma, Y.~Xu, and G.~Pan, ``Rsnn: Recurrent spiking neural networks for dynamic spatial-temporal information processing,'' in \emph{ACM Multimedia 2024}.

\bibitem{wu_spatio-temporal_2018}
Y.~Wu, L.~Deng, G.~Li, J.~Zhu, and L.~Shi, ``Spatio-{Temporal} {Backpropagation} for {Training} {High}-{Performance} {Spiking} {Neural} {Networks},'' \emph{Frontiers in Neuroscience}, vol.~12, p. 323875, 2018.

\bibitem{krizhevsky_cifar-10_2010}
A.~Krizhevsky, V.~Nair, and G.~Hinton, ``Cifar-10 (canadian institute for advanced research),'' vol.~5, no.~4, p.~1, 2010.

\bibitem{deng_imagenet_2009}
J.~Deng, W.~Dong, R.~Socher, L.-J. Li, K.~Li, and L.~Fei-Fei, ``{ImageNet}: {A} large-scale hierarchical image database,'' in \emph{Proceedings of CVPR}, Jun. 2009, pp. 248--255, iSSN: 1063-6919.

\bibitem{li_cifar10-dvs_2017}
H.~Li, H.~Liu, X.~Ji, G.~Li, and L.~Shi, ``{CIFAR10}-{DVS}: {An} {Event}-{Stream} {Dataset} for {Object} {Classification},'' \emph{Frontiers in Neuroscience}, vol.~11, 2017.

\bibitem{amir_low_2017}
A.~Amir, B.~Taba, D.~Berg, T.~Melano, J.~McKinstry, C.~Di~Nolfo, T.~Nayak, A.~Andreopoulos, G.~Garreau, M.~Mendoza, J.~Kusnitz, M.~Debole, S.~Esser, T.~Delbruck, M.~Flickner, and D.~Modha, ``A {Low} {Power}, {Fully} {Event}-{Based} {Gesture} {Recognition} {System},'' in \emph{Proceedings of {CVPR}}, 2017, pp. 7243--7252.

\bibitem{guo_recdis-snn_2022}
Y.~Guo, X.~Tong, Y.~Chen, L.~Zhang, X.~Liu, Z.~Ma, and X.~Huang, ``\BIBforeignlanguage{en}{{RecDis}-{SNN}: {Rectifying} {Membrane} {Potential} {Distribution} for {Directly} {Training} {Spiking} {Neural} {Networks}},'' in \emph{\BIBforeignlanguage{en}{Proceedings of {CVPR}}}, Jun. 2022, pp. 326--335.

\bibitem{yao_glif_2022}
X.~Yao, F.~Li, Z.~Mo, and J.~Cheng, ``\BIBforeignlanguage{en}{{GLIF}: {A} {Unified} {Gated} {Leaky} {Integrate}-and-{Fire} {Neuron} for {Spiking} {Neural} {Networks}},'' \emph{\BIBforeignlanguage{en}{Proceedings of NeurIPS}}, vol.~35, pp. 32\,160--32\,171, Dec. 2022.

\bibitem{deng_temporal_2021}
S.~Deng, Y.~Li, S.~Zhang, and S.~Gu, ``Temporal {Efficient} {Training} of {Spiking} {Neural} {Network} via {Gradient} {Re}-weighting,'' in \emph{Proceedings of {ICLR}}, Oct. 2021.

\bibitem{lian_learnable_2023}
S.~Lian, J.~Shen, Q.~Liu, Z.~Wang, R.~Yan, and H.~Tang, ``\BIBforeignlanguage{en}{Learnable {Surrogate} {Gradient} for {Direct} {Training} {Spiking} {Neural} {Networks}},'' in \emph{\BIBforeignlanguage{en}{Proceedings of {IJCAI}}}, Aug. 2023, pp. 3002--3010.

\bibitem{deng_tensor_2024}
H.~Deng, R.~Zhu, X.~Qiu, Y.~Duan, M.~Zhang, and L.-J. Deng, ``Tensor decomposition based attention module for spiking neural networks,'' \emph{Knowledge-Based Systems}, vol. 295, p. 111780, 2024.

\bibitem{lian_im-lif_2024}
S.~Lian, J.~Shen, Z.~Wang, and H.~Tang, ``{IM}-{LIF}: {Improved} {Neuronal} {Dynamics} {With} {Attention} {Mechanism} for {Direct} {Training} {Deep} {Spiking} {Neural} {Network},'' \emph{IEEE TETCI}, pp. 1--11, 2024.

\bibitem{wang_spatial-temporal_2023}
Y.~Wang, K.~Shi, C.~Lu, Y.~Liu, M.~Zhang, and H.~Qu, ``\BIBforeignlanguage{en}{Spatial-{Temporal} {Self}-{Attention} for {Asynchronous} {Spiking} {Neural} {Networks}},'' in \emph{\BIBforeignlanguage{en}{Proceedings of {IJCAI}}}, Aug. 2023, pp. 3085--3093.

\bibitem{yao_attention_2023}
M.~Yao, G.~Zhao, H.~Zhang, Y.~Hu, L.~Deng, Y.~Tian, B.~Xu, and G.~Li, ``Attention {Spiking} {Neural} {Networks},'' \emph{IEEE TPAMI}, vol.~45, no.~8, pp. 9393--9410, Aug. 2023.

\bibitem{yao_inherent_2023}
M.~Yao, J.~Hu, G.~Zhao, Y.~Wang, Z.~Zhang, B.~Xu, and G.~Li, ``\BIBforeignlanguage{en}{Inherent {Redundancy} in {Spiking} {Neural} {Networks}},'' in \emph{\BIBforeignlanguage{en}{Proceedings of the {ICCV}}}, 2023, pp. 16\,924--16\,934.

\bibitem{yao_temporal-wise_2021}
M.~Yao, H.~Gao, G.~Zhao, D.~Wang, Y.~Lin, Z.~Yang, and G.~Li, ``\BIBforeignlanguage{en}{Temporal-{Wise} {Attention} {Spiking} {Neural} {Networks} for {Event} {Streams} {Classification}},'' in \emph{\BIBforeignlanguage{en}{Proceedings of {ICCV}}}, 2021, pp. 10\,221--10\,230.

\bibitem{kundu_recent_2024}
S.~Kundu, R.-J. Zhu, A.~Jaiswal, and P.~A. Beerel, ``Recent {Advances} in {Scalable} {Energy}-{Efficient} and {Trustworthy} {Spiking} {Neural} {Networks}: from {Algorithms} to {Technology},'' in \emph{ICASSP}, Apr. 2024, pp. 13\,256--13\,260, iSSN: 2379-190X.

\end{thebibliography}

\end{document}